\documentclass[conference]{IEEEtran}
\IEEEoverridecommandlockouts
\usepackage{cite}
\usepackage{amsmath,amssymb,amsfonts}
\usepackage{graphicx}
\usepackage{textcomp}
\usepackage{xcolor}
\usepackage{algorithm}
\usepackage{algpseudocode}
\usepackage{cite}
\usepackage{amsmath,amssymb,amsfonts}
\usepackage{graphicx}
\usepackage{textcomp}
\usepackage[compatibility=false]{caption}

\usepackage[table]{xcolor}
\usepackage{adjustbox}
\usepackage{booktabs}
\usepackage{array}
\usepackage{tabularx}
\usepackage{multirow}
\usepackage{float}

\usepackage{tikz}
\usepackage{tcolorbox}

\usepackage{algorithmicx}
\usepackage{algpseudocode}
\usepackage{listings}
\newcommand{\myTitle}{RE-SciVQA}
\usepackage{xspace}

\definecolor{accessblue}{cmyk}{1,0.30,0,0.20}
\def\BibTeX{{\rm B\kern-.05em{\sc i\kern-.025em b}\kern-.08em
    T\kern-.1667em\lower.7ex\hbox{E}\kern-.125emX}}
\begin{document}

\title{RE-SciVQA: Improving Scientific VQA With Context-Relevant Retrieval and Lightweight MLLM Adaptation}

\author{\IEEEauthorblockN{1\textsuperscript{st} Sheikh Tahmim Mehdi}
\IEEEauthorblockA{\textit{Department of Electrical and Computer Engineering} \\
\textit{North South University}\\
Dhaka 1229, Bangladesh \\
sheikh.mehdi@northsouth.edu}
\and
\IEEEauthorblockN{2\textsuperscript{nd} Muminul Islam}
\IEEEauthorblockA{\textit{Department of Computer Science and Engineering} \\
\textit{United International University (UIU)}\\
Dhaka 1212, Bangladesh \\
mislam202115@bscse.uiu.ac.bd }
}

\maketitle

\begin{abstract}

A deep understanding of scientific literature often relies on interpreting figures that contain key information beyond the main text. Leveraging Large Language Models (LLMs) to extract insights from these figures can significantly accelerate knowledge discovery. Existing works on Scientific Visual Question Answering rely on resource-intensive training pipelines that are often inaccessible and require large computational resources. Instead of designing complex architectures, we propose RE-SciVQA, which focuses on improving the quality of retrieved context and fine-tuning small LLMs with parameter-efficient LoRA training. Using a simple query expansion step, relevant section headers are appended to the original question to improve the retrieval coverage. In addition, a simple adjustment of similarity scores is applied using figure references and caption alignment. On the SciVQA benchmark, RE-SciVQA achieves a ROUGE-1 F1 score of 73.9\%, outperforming chart-focused models such as UniChart (14.7\%), ChartGemma (6.9\%), and Matcha-ChartQA (6.3\%). Our relevance-enhanced RAG setup also improves the exact match by 11.9\% and the fuzzy match by 29.5\% compared with a standard RAG baseline, resulting in an overall gain of 14.3\%. We also examine closed-ended infinite visual questions, a category in which chart-understanding models often struggle. Our experiments show that using plot digitization instead of synthetic captions leads to an 18.5\% improvement for this question type.
\end{abstract}

\section{Introduction}


\IEEEPARstart{I}{n} a Visual Question Answering (VQA) system~\cite{sima2024drivelm, ishmam2024image, tanaka2023slidevqa}, a text-based question and an image are provided to a vision-language model (VLM) to help it extract and interpret relevant information from the visual content~\cite{antol2015vqa}. The image is often paired with a caption, which enhances the model's ability to understand the visual context by integrating multiple modalities. Scientific Visual Question Answering (SciVQA) involves answering questions based on figures from scientific papers~\cite{borisova2025scivqa, pramanick2024spiqa, li2023scigraphqa}. In this context, Multimodal Large Language Models (MLLMs) are given a scientific figure, its caption, and a question. The model's task is to select the correct answer from a set of predefined, closed-ended options.



In recent years, many studies have focused on Chart Question Answering (ChartQA)~\cite{hoque2022chart, masry2022chartqa, wu2024chartinsights}. However, understanding scientific visual questions remains relatively underexplored ~\cite{methani2020plotqa, masry2022chartqa}. In scientific papers, the main idea is typically demonstrated through visuals such as charts, figures, and diagrams. The information in these scientific papers is often spread across different sections of a paper~\cite{head2021augmenting}. As a result, tasks in the scientific domain require precise numerical reasoning and the ability to interpret complex visual layouts alongside detailed textual descriptions.

Recently, there has been a clear shift towards smaller and more efficient Multimodal Large Language Models (MLLMs) ~\cite{xu2024chartmoe, zhang2024tinychart}, such as TinyChart (3B) and ChartLlama~\cite{zhang2025mllms}. This trend is driven by the high computational cost and resource demands of very large models (often exceeding 70B parameters). However, using smaller models typically comes with a performance trade-off, as they still lag behind larger models in accuracy. Moreover, when it comes to scientific diagrams, even with the latest model and a large dataset, researchers are still struggling~\cite{borisova2025scivqa}.

LLM visualizes a chart or a graph as an image patch, sometimes it can not find the bar that is located at exact coordinates corresponding to the label on its \texttt{x} and \texttt{y} axes~\cite{zhang2025mllms}. Moreover, MLLM often guesses while not calculating the actual result. Thus, previous work, such as TinyChart~\cite{zhang2024tinychart} demonstrated ``Program-of-Thought'' instead of ``Chain-of-Thought'' to solve this. Also, for most scientific charts, we have images but not the sources (i.e., Excel/CSV files) to generate the image~\cite{masry2025bigcharts}. That also makes it difficult to get accurate results.

\begin{figure}
\centering
\includegraphics[width=0.85\linewidth]{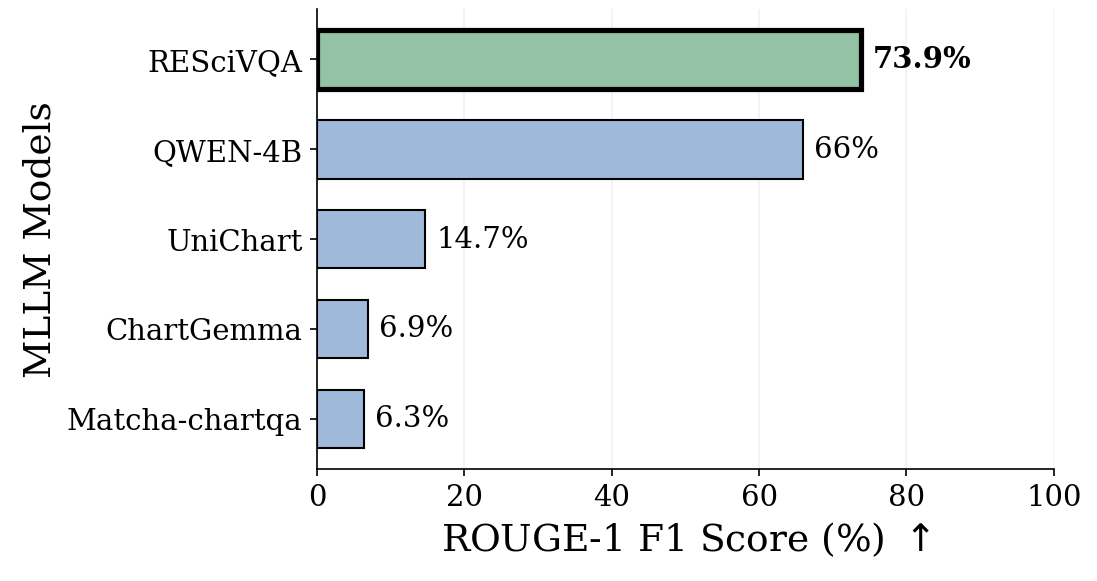}
\caption{Comparative analysis of model performance on the \myTitle benchmark. The chart plots the performance of various models based on the ROUGE-1 F1 score~\cite{lin-2004-rouge} (\%). \myTitle achieves a peak performance of 73.9\%, significantly outperforming other evaluated models such as UniChart (14.7\%), ChartGemma (6.9\%), and Matcha-ChartQA (6.3\%).}
\label{fig:intro}
\end{figure}

In addition, recent research reveals that Retrieval-Augmented Generation (RAG) is found to be effective for enhancing static models. Bonomo et al.~\cite{bonomo2025visual} proposed Visual RAG, compared to the most recent state of the art which can obtain an accuracy higher while using a much smaller set of demonstrating examples. In addition, we can see the RAG-grounded VLM performed better, whereas the same VLM without retrieval performed worse in both accuracy and precision~\cite{zhang2025seeing}. Bhat et al.~\cite{bhat2025expertneurons} introduced RAVQA-VLM, which achieved significant improvements targeting closed-ended visual, nonvisual questions over scientific figures and  efficient computing which gave us significant proof of RAG's potential for Scientific Visual Question Answering (SciVQA).

In this study, we propose an architecture \myTitle that achieves a good accuracy for SciVQA with a resource effective model. We adopt a retrieval centric architecture to maximize our performance with fine-tuning Qwen-3 4B with QLoRA. Figure \ref{fig:intro} presents the performance comparison between \myTitle and other MLLMS.  Furthermore, we carried out experiments to demonstrate whether visual grounding could be alleviated by captioning through ChartGemma and Deplot. The key contributions of this study are listed as follows: 
\begin{enumerate}


    \item We propose \myTitle,
which focuses on improving the quality of retrieved context and fine-tuning small LLMs with parameter-efficient LoRA training.
    \item \myTitle (4B) outperforms smaller models such as ChartGemma (R-1 F1 = 0.069) and Matcha-Chartqa (R-1 F1 = 0.063), improving their performance by 10.1x and 11.7x, respectively. This demonstrates the effectiveness of our retrieval-centric fine-tuning approach.

    \item \myTitle (4B) model achieves an R-1 F1 score of 0.739 and R-L F1 of 0.738, which is comparable to larger models such as InternVL3 (8B) (R-1 F1 = 0.740) and Qwen2.5-VL (7B) (R-1 F1 = 0.695), despite being much smaller.

\end{enumerate}

\section{Related Work}

Recent developments in chart based question answering pivoted towards instruction tuning on diverse chart based datasets to enhance interpretation of charts. In particular, ChartGemma~\cite{masry2025chartgemma} achieved competitive performance by integrating a specialized encoder along with an LLM based decoder. However, these chart understanding models often exhibit good performance on benchmark datasets, but their performance degrades in the scientific domain as  shown in Figure \ref{fig:intro}.  

To address these domain specific challenges, prior work on Scientific Visual Question Answering involved the use of advance prompting techniques such as chain of thought alongside designing an ensemble technique to select the most suitable VLM for figure type~\cite{movva2025enhancing}. However, this prompting technique is resource intensive and also results in inference latency.

Integrating Retrieval-Augmented Generation (RAG) with Vision-Language Models is a promising direction for handling knowledge-intensive tasks. Bhat et al.~\cite{bhat2025expertneurons} successfully combined a fine-tuned InternVL-3-14B model with dense retrieval to answer scientific questions. However, Bhat et al. reported that their RAG system still suffered from hallucinations due to the retrieval of irrelevant or insufficient context. A critical limitation of this retrieval mechanism is the reliance on regular expression (Regex) to parse the scientific paper. This text based extraction ignores the variance in scientific layouts.

Preserving the semantic structure of scientific documents is a prerequisite for effective multimodal reasoning. Pfitzmann et al.~\cite{pfitzmann2022doclaynet} argue that accurate layout analysis is critical, as traditional parsers often disrupt the logical flow between text and figures~\cite{shen2022vila, nassar2022tableformer}. To address this, recent visual-layout detection models such as DocLayNet~\cite{pfitzmann2022doclaynet} have allowed precise identification of elements such as headers, captions, and figures. To address the need for structured inputs, we utilize Docling~\cite{auer2024docling, livathinos2025docling}, an open-source toolkit built upon DocLayNet, which enables us to parse different formats of documents to a specific concise layout efficiently.

Beyond structural accuracy, it is also important to consider efficiency. Most of the current research on the scientific domain conducted with massive multimodal backbones, for instance, ~\cite{bhat2025expertneurons} used InternVL3 with 14B parameters. These architectures are resource intensive and remain computationally prohibitive for many academic and industrial applications. 

To address these constraints, our framework utilizes a compact Qwen3-4B model. However, aligning this model with the scientific domain requires Supervised Fine-tuning (SFT). A recent study~\cite{yang2024quantized} shows that Quantized Low Rank Adaptation (QLoRA)~\cite{dettmers2023qlora}is effective for VQA task, allowing the compact model to achieve competitive reasoning abilities by freezing the pretrained weights in a compressed 4-bit  format while training other smaller adapter layers. By utilizing this technique, we have adapted the model to the scientific domain while maintaining computational efficiency.

\section{Methodology}


In this section, we present \myTitle, a multi-stage framework designed for  scientific figure understanding. Our approach consists of three key stages: (a) preprocessing step, (b) context-aware retrieval, and (c) parameter-efficient LoRA fine-tuning. The algorithm of \myTitle  is demonstrated in Algorithm~\ref{alg:training}.

\begin{algorithm}[!htpb]
\caption{\myTitle: Retrieval-Aware QLoRA Training}
\label{alg:training}
\begin{algorithmic}[1]

\State \textbf{Phase 2: Metadata-Aware Re-ranking}
\For{candidate $c_i \in \mathcal{C}_{pool}$}
    \State $S_{\text{meta}}(c_i) \leftarrow \sum_{k \in \mathcal{K}} \beta_k \cdot \mathbb{I}_k(c_i)$ \Comment{Calculate metadata bonuses}
    \State $S_{\text{final}}(c_i) \leftarrow S_{\text{rel}}(c_i) + S_{\text{meta}}(c_i)$ \Comment{Combine scores}
\EndFor

\State $\mathcal{C}_{\text{best}} \leftarrow \underset{c_i \in \mathcal{C}_{pool}}{\text{ArgTop-}K} \{ S_{\text{final}}(c_i) \}$ \Comment{Final Context Selection}

\Statex \textbf{Phase 3: Parameter-Efficient Optimization (QLoRA)}
\State $W_0^{4\text{bit}} \leftarrow Q_{\text{NF4}}(W_0)$ \Comment{Freeze Base Model in 4-bit}
\State Initialize $\theta_{\text{LoRA}} = \{A, B\}$ where $A \sim \mathcal{N}(0, \sigma^2), B = 0$
\State $\mathcal{L}(\theta) \leftarrow -\sum \log P(A_b \mid P_b; \theta)$ \Comment{Cross-Entropy Loss for batch}

\While{not converged}
    \For{batch $(P_b, A_b) \in \mathcal{D}_{\text{instruct}}$}
        \State $h \leftarrow W_0^{4\text{bit}} P_b + \frac{\alpha}{r} (BA) P_b$ \Comment{Forward Pass with Adapters}
        \State $loss \leftarrow \mathcal{L}(h, A_b)$
        \State Update $\theta_{\text{LoRA}} \leftarrow \theta_{\text{LoRA}} - \eta \nabla_{\theta_{\text{LoRA}}} loss$ \Comment{Backprop only on Adapters}
    \EndFor
\EndWhile

\State \Return $\theta_{\text{LoRA}}$
\end{algorithmic}
\end{algorithm}

\begin{table}[h]
\caption{Summary of key notations used in the \myTitle framework.}
\centering
\renewcommand{\arraystretch}{1.3} 

\label{tab:notations}
\begin{tabularx}{\columnwidth}{@{} l X @{}} 
\toprule
\textbf{Notation} & \textbf{Description} \\
\midrule

$q, q', Q_{\exp}$ & Original query, single expanded query, and the full set of expanded queries. \\
$\Phi(\cdot), \mathbf{V}$ & Text embedding function and the pre-computed document block embeddings. \\
$\mathcal{C}_{pool}, \mathcal{C}_{\text{best}}$ & Initial retrieved candidate pool and the final selected top-$K$ context blocks. \\
$N, K$ & Count of initially retrieved blocks ($N$) and final selected blocks ($K$). \\
$c_i$ & The $i$-th candidate text block in the retrieval pool. \\
$S_{\text{rel}}(c_i)$ & Cosine similarity score between the expanded query and candidate $c_i$. \\
$S_{\text{meta}}(c_i)$ & Metadata-based heuristic bonus score for candidate $c_i$. \\
$S_{\text{final}}(c_i)$ & Final re-ranked score (combining similarity and metadata bonuses). \\
$\mathcal{K}, \beta_k, \mathbb{I}_k$ & Set of metadata attributes, their scalar weights, and the indicator function. \\

$W_0, W_0^{4\text{bit}}$ & Pre-trained base weight matrix and its 4-bit quantized (NF4) state. \\
$A, B$ & LoRA trainable low-rank decomposition matrices. \\
$r, \alpha$ & LoRA rank (intrinsic dimension) and the adapter scaling factor. \\
$\theta_{\text{LoRA}}$ & Complete set of trainable parameters within the LoRA adapter. \\
$P_b, A_b$ & Formatted prompt (input context) and answer (ground truth) for a training batch. \\

\bottomrule
\end{tabularx}

\end{table}

\subsection{Pre-processing}
\subsubsection{Image Up-scaling}
Influence of image quality is observed on the performance of multimodal large language models (MLLMs)~\cite{lu2023empirical}. Prior work has shown that increasing image resolution can significantly improve MLLM performance~\cite{lu2023empirical}. Motivated by these findings, we enhance the visual clarity of the figures by applying Lanczos interpolation for image upscaling.

\subsubsection{Parsing}


We employ a custom parsing pipeline built on Docling~\cite{livathinos2025docling} to extract structured text blocks from scientific papers. The parsed content, along with relevant metadata, is stored in JSON format and later used as retrieval units in our RAG framework. Each paper is first processed by Docling’s document converter, which extracts the main document body and associated section headers. We define a set of target semantic types such as abstract, introduction, caption, and related sections to guide the parsing process. The extracted text blocks are normalized and then classified into one of these predefined semantic categories before being saved accordingly.


\begin{figure*}
\centering
\includegraphics[width=0.95\linewidth]{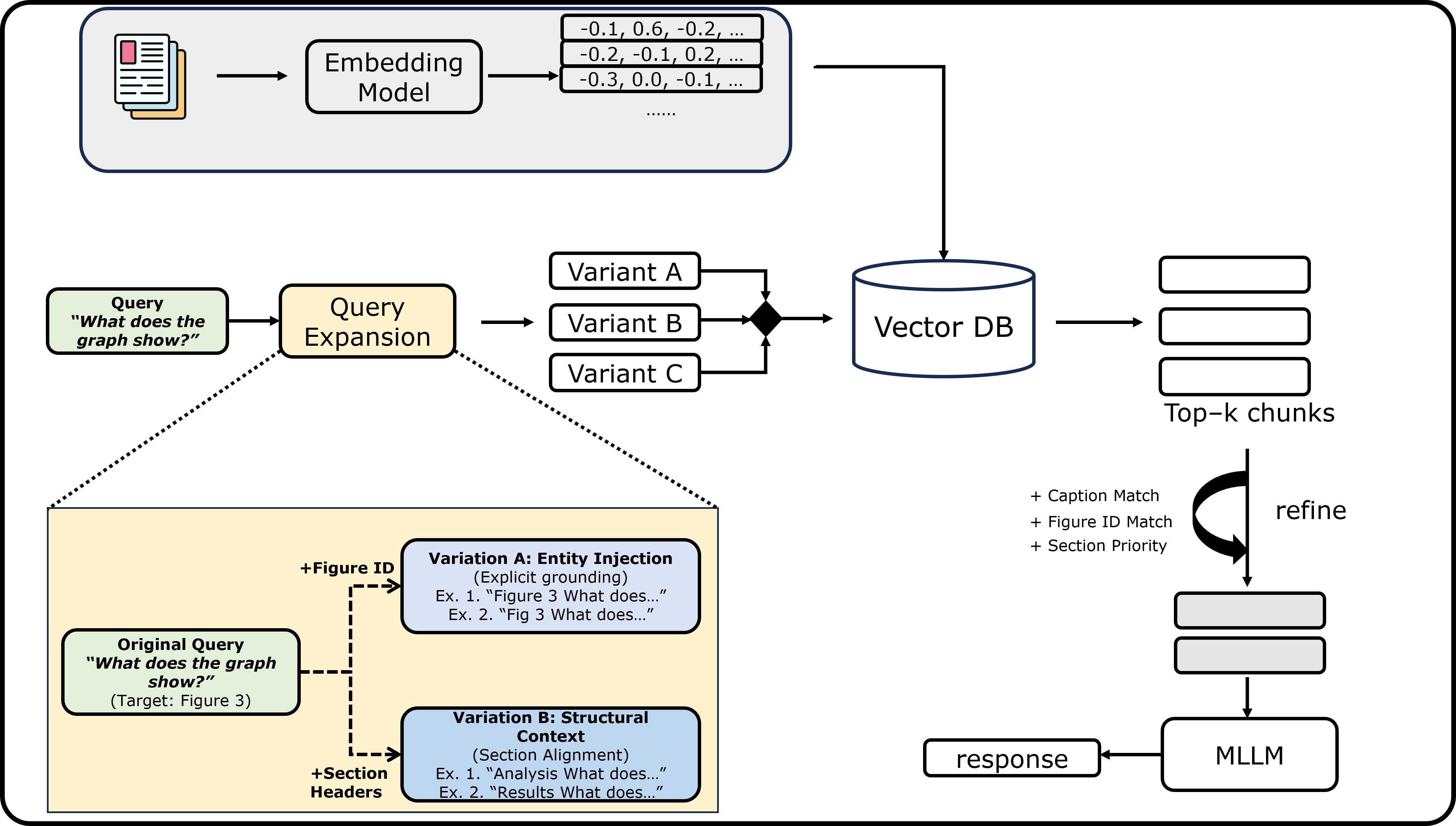}
\caption{\myTitle Architecture diagram}
\label{fig:rag_d}
\end{figure*}

\subsection{Context Retrieval - Phase 1}
The primary goal of RAG is to retrieve relevant text blocks from the parsed paper to enhance contextual sufficiency when answering questions. This enables the MLLM to better interpret figures when the visual content and caption alone do not provide sufficient information. To achieve this, we embed both the question and the caption, then locate the most relevant text blocks through cosine similarity search, as illustrated in the Figure~\ref{fig:rag_d}.





To maximize context, the two top relevant blocks are selected. Initially, we find the vector representation of the parsed blocks with an embedding model. The figure number is appended to the initial question to ensure that the retrieval system focuses on the text blocks that refer to this specific figure. we also concatenate section names in the query; for instance, a section header like `analysis' is concatenated with the query in case there are keywords like `show' or `display' in the question. This expansion would result in an enriched semantic representation, increasing the possibility of finding the correct block. The query expansion strategy is illustrated in Figure \ref{fig:rag_d}.

The retrieval process consists of two stages: candidate pooling and re-ranking among candidates~\cite{glass2022re2g}.
In the first stage, a candidate pool $\mathcal{C}$ is generated by retrieving the top-$N$ most relevant blocks for each query in the expanded query set using cosine similarity search.
\begin{equation}\label{candidate-pooling}
\mathcal{C} = \bigcup_{q' \in Q_{\exp}} \text{Top-}N \left( \Phi(q') \cdot \mathbf{V} \right)
\end{equation}
where $\Phi(q')$ represents the query embedding and $\mathbf{V}$ denotes the document block embeddings. It is to be noted that the notations are demonstrated in Table~\ref{tab:notations}.

In the second stage, candidates are re-ranked using a light weight heuristic scoring system that combines similarity scores with metadata-based bonuses, and the top-$K$ candidates are selected. 
\begin{equation}\label{scoring-system}
S_{\text{final}}(c_i) = S_{\text{rel}}(c_i) + \sum_{k \in \mathcal{K}} \beta_k \cdot \mathbb{I}_k(c_i)
\end{equation}
Here, $S_{\text{rel}}(c_i)$ denotes the cosine similarity score for the $i$-th candidate from the pool, $\mathbb{I}_k(c_i)$ is an indicator function that equals 1 if candidate $c_i$ contains the $k$-th metadata attribute (e.g., figure matching or specific section type), and 0 otherwise, and $\beta_k$ is the scalar weight assigned to the $k$-th metadata attribute where we added bonus score for the figure ID match +0.15, caption match +0.10 and section priority +0.08 for enhancing relevance. The re-ranking weights were heuristically selected. The Figure ID match has the dominant weight due to the direct figure reference, which indicates strong relevance, followed by the caption overlap and section priority. We have tried to find the optimal weight through grid search experimentation; however, the resulted weight lead to negligible improvement in retrieval. Therefore, we retained the heuristic weights to maintain a light-weight computationally efficient re-ranking.

Finally, we select the top-$K$ candidates based on $S_{\text{final}}(c_i)$. This scoring system allows us to perform re-ranking without an additional model and helps us achieve a resource-efficient architecture.

\begin{equation}\label{final-context-selection}
P_{\text{context}} = \underset{c_i \in \mathcal{C}_p}{\text{ArgTop-}K}
\left\{S_{\text{final}}(c_i)\right\}.
\end{equation}


\subsection{LoRA Finetuning}
 Low Rank Adaptation (LoRA)~\cite{hu2022lora} refers to training only a small number of new matrices by freezing the remaining weights. We used Quantized Low-Rank Adaptation (QLoRA)~\cite{dettmers2023qlora} to fine-tune the 4-billion parameter model for the task of SciVQA to avoid the high cost of standard fine-tuning.  To reduce computational overhead during training, a JSONL file is compiled offline with all necessary data, like retrieved context using RAG, question, caption, ground truth and image path. This method freezes the pre-trained model weights and injects trainable rank decomposition matrices into each layer of the Transformer architecture.

To fit the training process within limited VRAM constraints while maintaining gradients, the base model $W_0$ was loaded using the NF4 (NormalFloat 4-bit) data type, a theoretically optimal quantization for normally distributed weights and also used Double Quantization to save an additional 0.4 bits per parameter on average.


\subsubsection{Mathematical Formulation}

Following the QLoRA formulation proposed by Hu et al.~\cite{hu2022lora}, for a pre-trained weight matrix $W_0 \in \mathbb{R}^{d \times k}$, the modified forward pass for a linear layer mapping input $P_b$ to output $h$ is defined as:

\begin{equation}
    h = W_0 P_b + \frac{\alpha}{r} (BA)P_b
\end{equation}

Here, $W_0$ remains frozen (4-bit quantized), $B$ is initialized to zeros, and $A$ is initialized with a random Gaussian distribution, ensuring $\Delta W = 0$ at the start of training, $\frac{\alpha}{r}$ is a scaling factor that regulates the magnitude of the adapter's influence.

\subsubsection{Target Module Selection}
We froze the vision encoder parameters to preserve the model's pre-trained visual grounding capabilities. The QLoRA adapters were injected only into the LLM.

Standard implementations often target only the query ($Q$) and value ($V$) projections but we applied adapters to all linear layers within the Transformer blocks. The specific target modules were the attention modules, feed-forward modules, and additional three projection specific to SwigLU. 



\subsection{Prompt Formulation}
To effectively ground the MLLM in the scientific context, we employ a structured, dynamic prompting strategy. The prompt is constructed by concatenating distinct semantic blocks as System Instructions, Visual Grounding, Retrieved Context, Task and Output Constraints, depending on the data available for each sample.

\noindent The prompt construction follows the template illustrated here:

\begin{figure}
    \centering
    \begin{tcolorbox}[colback=gray!5, colframe=black!75, arc=2mm, title=\textbf{Prompt}]
        \small
        \textbf{[System Instruction]} \\
        You are a scientific visual question answering assistant. Provide direct answers without showing your reasoning process.
        
        \vspace{0.2cm}
        \textbf{[Visual Grounding]} \\
        Figure Caption: \texttt{<caption\_text>}
        
        \vspace{0.2cm}
        \textbf{[Retrieved Context]} \\
        Context: \texttt{<caption\_context>} \\
        Additional context: \texttt{<question\_context>}
        
        \vspace{0.2cm}
        \textbf{[Task]} \\
        Question: \texttt{<question\_text>}
        
        \vspace{0.2cm}
        \textbf{[Output Constraints]} \\
        \textit{Condition: Multiple Choice} \\
        Options: \texttt{<options\_list>}. Answer with ONLY the letter (A/B/C/D). Do not show your thinking process. \\
        Provide a direct answer only. Do not show your thinking process.
    \end{tcolorbox}
    \caption{Structure of the prompt used in \myTitle.}
    \label{fig:prompt}
\end{figure}






\textbf{System Instruction:} We explicitly define the model's role as a ``scientific visual question answering assistant'' to condition it for domain-specific reasoning. Crucially, we enforce a strict negative constraint (``without showing your reasoning process'') to prevent the model from generating verbose Chain-of-Thought (CoT) explanations, which simplifies parsing the final answer.

\textbf{Adaptive Output Constraints:} The tail of the prompt adapts to the question types. For multiple choice questions, we append the option set and enforce a single-token constraint (``Answer with ONLY the letter'') to ensure valid evaluation against ground truth. For open-ended questions, we reinforce the ``direct answer only'' constraint to mitigate hallucinated conversational filler.

\section{Results}

\subsection{Datasets}

\begin{table}[!ht]
    \centering
     \caption{Train, validation and test split for the SciVQA dataset.} 
    \label{tab:dataset_stats}
    \begin{tabular}{|c|c|}
        \hline
        \textbf{Dataset } & \textbf{Number of Samples} \\
        \hline
        Train & 15,100 \\
        Validation & 1,680 \\
        Test & 4,200 \\
        \hline
        \textbf{Total} & \textbf{20,980} \\
        \hline
    \end{tabular}
\end{table}
The SciVQA dataset~\cite{borisova-etal-2025-scivqa} consists of scientific figures from arXiv and ACL. The dataset is split into 3 parts: train, test and validation. Each figure in the dataset is associated with 7 questions. The dataset contains seven different types of questions both visual and non-visual. The description of dataset is shown in Table~\ref{tab:dataset_stats}.  

\subsection{Implementation Details}

Due to resource constraint, we used a small model named Qwen3-VL-4B-Instruct. We choose the Docling~\cite{livathinos2025docling} parser to parse scientific papers, These parsed blocks were then embedded using E5-Large-V2 model to create FAISS indexing~\cite{douze2025faiss} for the vector database.   

Moreover, we have finetuned Qwen 3 using QLoRa~\cite{dettmers2023qlora} with rank 32 along with alpha 64. The training was conducted with  a learning rate of $2 \times 10^{-5}$ with a cosine scheduler. The loss converged within a single epoch, and all these experiments were conducted in Google cloud.

\begin{table*}[!htpb]
\caption{Performance comparison of \myTitle vs other models  with various sizes, comprising with and without Chain-of-Thought (CoT) on SciVQA dataset.}
\label{tab:performance_comparison}
\centering
\resizebox{0.6\linewidth}{!}{
\begin{tabular}{|l|c|c|c|}
\hline
 \textbf{Model} & \textbf{Size} & \textbf{R-1 F1} & \textbf{R-L F1} \\
\hline

\myTitle & \textbf{4B} & \textbf{0.739} & \textbf{0.738} \\

ChartGemma & 3B  & 0.069 & 0.066 \\

Matcha-Chartqa & 282M & 0.063 & 0.063 \\

UniChart & 140M & 0.147 & 0.147 \\
\hline
\multicolumn{4}{|c|}{\cellcolor{teal!20}\textbf{With Chain of Thought (CoT)}} \\
\hline
InternVL3~\cite{movva2025enhancing} & 8B & 0.740 & 0.740 \\
Qwen2.5-VL~\cite{movva2025enhancing} & 7B & 0.695 & 0.694 \\
Bespoke~\cite{movva2025enhancing} & 7B & 0.709 & 0.708 \\
Phi-4~\cite{movva2025enhancing} & 5.4B & 0.562 & 0.561 \\

\hline
\multicolumn{4}{|c|}{\cellcolor{cyan!20}\textbf{Without Chain of Thought (CoT)}} \\
\hline
InternVL3~\cite{movva2025enhancing} & 8B & 0.727 & 0.727 \\
Qwen2.5-VL~\cite{movva2025enhancing} & 7B & 0.633 & 0.633 \\
Bespoke~\cite{movva2025enhancing} & 7B &  0.652 &  0.651 \\
Phi-4~\cite{movva2025enhancing} & 5.4B & 0.544 &  0.543 \\
\hline
\multicolumn{4}{|c|}{\cellcolor{brown!20}\textbf{Large Model}} \\
\hline
InternVL3 LORA Fine-tuned + RAG ~\cite{bhat2025expertneurons} & \textbf{14B} & \textbf{0.798} & \textbf{0.798} \\
\hline

\hline
\end{tabular}}
\end{table*}

\subsection{Efficiency Analysis with varying Model Size}
To analyze efficiency with model size, we benchmark our method with various model sizes ranging from 140M to 14B as shown in Table \ref{tab:performance_comparison} in SciVQA dataset. We use the ROUGE-1 F1~\cite{lin-2004-rouge} and ROUGE-L F1~\cite{lin-2004-rouge} score as our primary evaluation metrics to measure the ability for scientific reasoning.


To find the capabilities of existing compact models, we evaluated Matcha-ChartQA (282M) and UniChart (140M). As shown in Table \ref{tab:performance_comparison}, we observe that \myTitle substantially dominates these small models by a large margin. 

ChartGemma 3B has similarly exhibited weak scientific reasoning, highlighting a substantial performance disparity (R1-F1 0.069) relative to \myTitle, even though it is a 3B-parameter model.



Remarkably, \myTitle achieves near-parity with the top-performing InternVL3 8B, trailing by a negligible margin of merely $0.13\%$. This performance is achieved despite our model having $50\%$ fewer parameters and without the computationally expensive Chain-of-Thought (CoT) reasoning.

In the CoT included configurations, \myTitle remains competitive despite not using explicit reasoning chains. It outperforms Qwen2.5-VL (7B) by 6.3\%, Bespoke (7B) by 4.2\%, and Phi-4 (5.4B) by 31.5\%. Excluding CoT, the advantage of \myTitle becomes even more evident. It achieved accuracy gains of $1.7\%$ over InternVL3, $16.7\%$ over Qwen 2.5 VL, $13.3\%$ over Bespoke, and $35.8\%$ over Phi-4.

We also compared \myTitle with a model Bhat et al.~\cite{bhat2025expertneurons} employed, a configuration closely aligned to our approach but with approximately 3.5 times more parameters. Despite this substantial increase in model capacity, the performance of  \myTitle is trailed by only 7.4\%, highlighting its strong efficiency.

\subsection{Result Analysis}
Table~\ref{tab:final_comprehensive_comparison} compares the zero-shot performance of the Qwen-3 4B model against our proposed architecture, \myTitle, which is fine-tuned using QLoRA with RAG context. 

For the zero-shot setup, we evaluate the model using image captions, achieving an R1-F1 score of 0.660. To test whether contextual sufficiency would improve performance, we use the RAG context, which eventually degrades performance by 11.7\%. Here the base model treats the RAG context as a distraction, leading to increased hallucinations.

We infer \myTitle without RAG to find out the RAG context's effect on the fine-tuned model. \myTitle(without RAG) results in a notable improvement of 10.9\% over the baseline, raising the score of R1-F1  to 0.732. Moreover, we achieve peak performance in \myTitle(with RAG) improving 12\% over the baseline with a score of R1-F1 0.739.

\begin{table*}
\centering
\scriptsize
\caption{ \myTitle vs.  Qwen-3 4B Zero-Shot performace comparison across different question types using R1-F1 and RL-F1 Metrics.}
\label{tab:final_comprehensive_comparison}
\resizebox{\textwidth}{!}{
\begin{tabular}{|l|c|c|c|c|c|c|c|c|}
\hline
{\textbf{Question Type}} & \multicolumn{4}{c|}{\textbf{Qwen-3 4B (Zero-Shot + caption)}} & \multicolumn{4}{c|}{\textbf{\myTitle}} \\ \cmidrule{2-9} 
 & \multicolumn{2}{c|}{\textbf{Baseline}} & \multicolumn{2}{c|}{\textbf{with RAG}} & \multicolumn{2}{c|}{\textbf{without RAG}} & \multicolumn{2}{c|}{\textbf{with RAG}} \\ \cmidrule{2-9} 
 & \textbf{R1-F1} & \textbf{RL-F1} & \textbf{R1-F1} & \textbf{RL-F1} & \textbf{R1-F1} & \textbf{RL-F1} & \textbf{R1-F1} & \textbf{RL-F1} \\ \hline

Closed-ended infinite (visual) & 0.488 & 0.488 & 0.497 & 0.496 & 0.518 & 0.517 & 0.541 & 0.541 \\
Closed-ended infinite (non-visual) & 0.655 & 0.653 & 0.647 & 0.643 & 0.672 & 0.669 & 0.676 & 0.673 \\
Closed-ended finite (binary visual) & 0.735 & 0.735 & 0.736 & 0.736 & 0.738 & 0.738 & 0.771 & 0.771 \\
Closed-ended finite (binary non-visual) & 0.761 & 0.761 & 0.761 & 0.761 & 0.793 & 0.793 & 0.821 & 0.821 \\
Closed-ended finite (non-binary visual) & 0.679 & 0.679 & 0.676 & 0.676 & 0.727 & 0.727 & 0.736 & 0.736 \\
Closed-ended finite (non-binary non-visual) & 0.644 & 0.644 & 0.669 & 0.669 & 0.762 & 0.762 & 0.718 & 0.718 \\ 
Unanswerable & 0.661 & 0.658 & 0.093 & 0.079 & 0.915 & 0.915 & 0.908 & 0.908 \\ \hline
\textbf{Overall} & \textbf{0.660} & \textbf{0.660} & \textbf{0.583} & \textbf{0.580} & \textbf{0.732} & \textbf{0.732} & \textbf{0.739} & \textbf{0.738} \\ \hline 
\end{tabular}}
\end{table*}

The introduction of the RAG context to the model with zero-shot configuration results in an overall performance decrease of approximately $12\%$ in the R1-F1 metric. As evidenced in Table~\ref{tab:final_comprehensive_comparison}, the most significant degradation occurs within the unanswerable question type. The R1-F1 score drops by 86\% and RL-F1 by 88\%. This substantial decrease in performance can be attributed to contextual noise~\cite{yang-etal-2025-llm-reasoning} or distraction effect~\cite{shi2023large}, where the model attempts to derive an answer from irrelevant RAG Context rather than correctly identifying the question as unanswerable. For closed-ended finite (binary non-visual), for both R1 and RL score 0.761, with and without RAG context, implies that the RAG context does not have any significance for binary answers. We observe some minor change in other question type.

\myTitle achieves R1-F1 score improvement over our baseline in closed-ended infinite (visual) by 10.9\%,  closed-ended infinite (non-visual) by 3.2\%, closed-ended finite (binary visual) by 4.8\%, closed-ended finite (binary non-visual) by 7.9\%, closed-ended finite (non-binary visual) by 8.4\%, closed-ended finite (non-binary non-visual by 18.3\% and unanswerable by 38.4\%. 

We further observe R1-F1 improvement of \myTitle with RAG over \myTitle without RAG in closed-ended infinite (visual) by 4.4\%,  closed-ended infinite (non-visual) by 0.59\%, closed-ended finite (binary visual) by 4.5\%, closed-ended finite (binary non-visual) by 3.5\%, closed-ended finite (non-binary visual) by 1.2\%.

But in closed-ended finite (non-binary non-visual) and unanswerable type question without RAG context \myTitle without RAG achieves 6.1\% and 0.7\% improvement respectively over \myTitle with RAG.

In \myTitle with RAG, a slight drop in unanswerable questions proves that even a trained model try to answer from the RAG context in the absence of an answer. In conclusion, the RAG context can hinder the performance of a base model due to increased hallucinations. Conversely, the \myTitle approach demonstrates that fine-tuning by integrating RAG context can enhance performance by leveraging the context and distinguishing noise.
\subsection{Qualitative Comparison}
\begin{figure}
\centering
\includegraphics[width=0.9\linewidth]{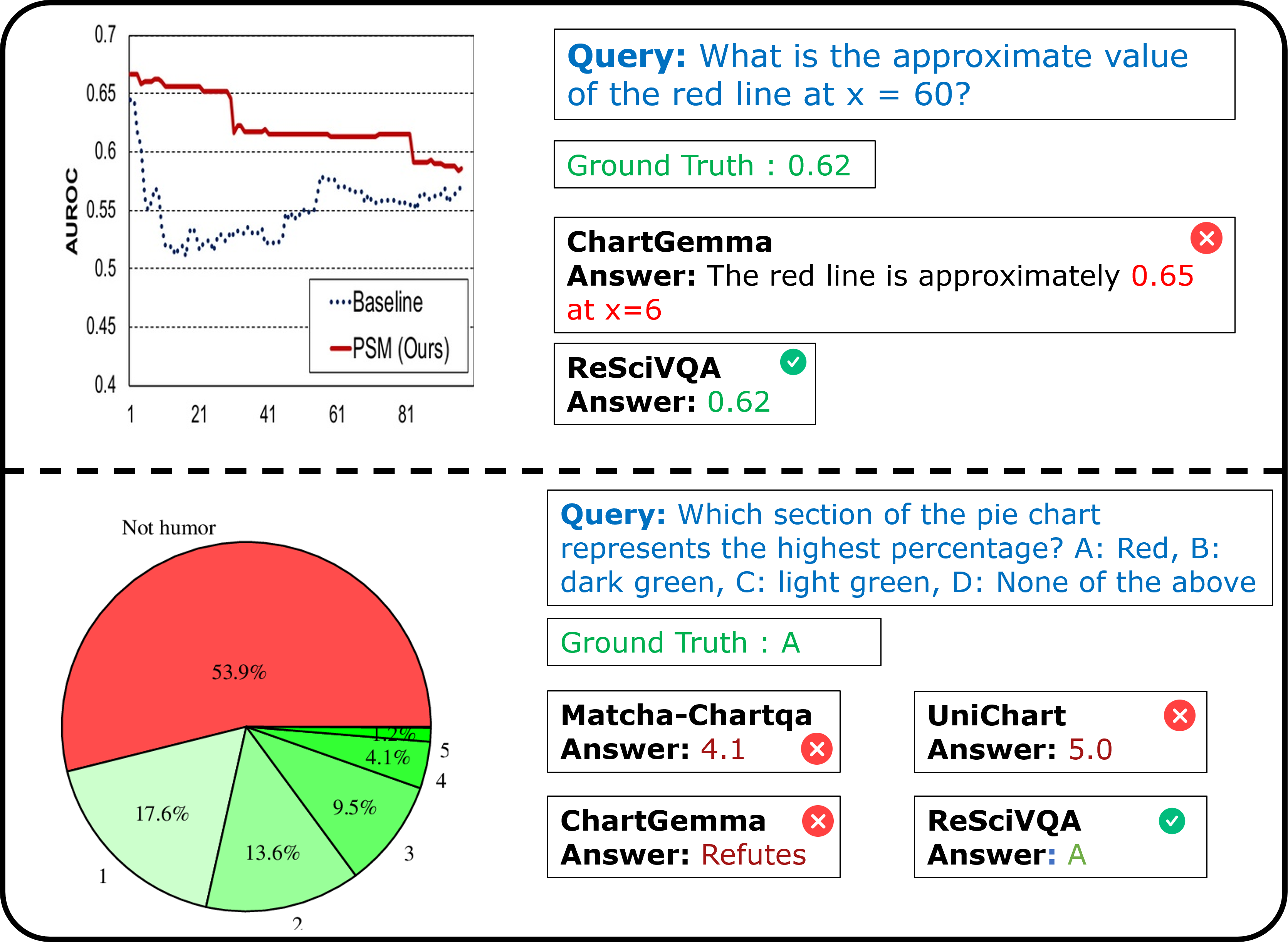}

\includegraphics[width=0.9\linewidth]{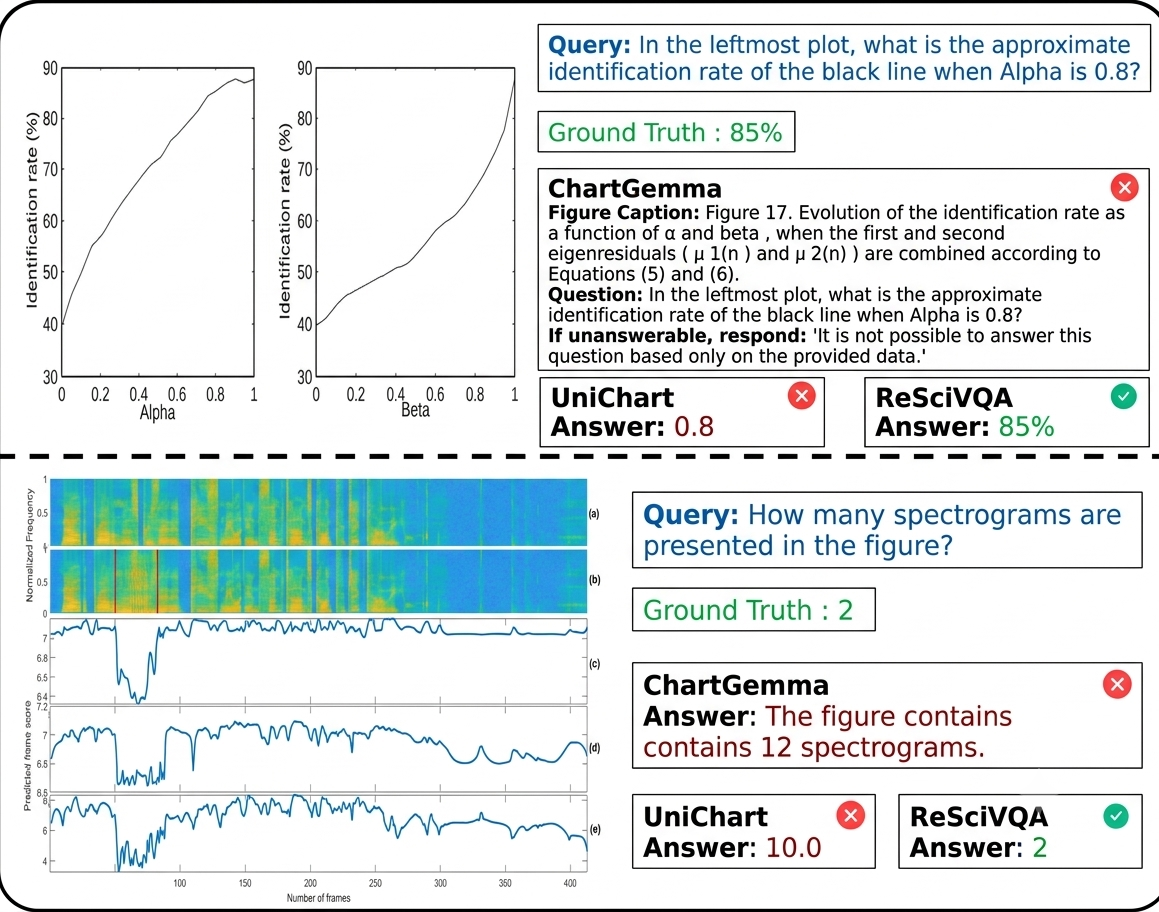}
\caption{Qualitative comparison of \myTitle with other baselines.}
\label{fig:chartgemma-vs-our-approach}
\end{figure}

We demonstrate the quality of response generation by \myTitle in Figure~\ref{fig:chartgemma-vs-our-approach}. 

The first example shows the comparison between \myTitle and ChartGemma, as demonstrated by correctly estimating the red line's value at x=60 (0.62), whereas the baseline model ChartGemma returned an incorrect value of 0.65. In contrast, in the second example, while other baselines succeed in identifying the sections, e.g., identifying the red section as the highest percentage, they often struggle with output format, providing numerical values (like 4.1 or 5.0) instead of the correct categorical options (A, B, C, D). 

\begin{table}
\centering
\caption{Comparison of retrieval quality between baseline RAG vs our proposed retrieval mechanism with heuristic scoring along with query expansion. Answer presence measures the frequency of the ground truth answer string appearing within the retrieved context.}
\begin{tabular}{l c c}
\toprule
\textbf{Metric} &  \textbf{Baseline RAG} & \textbf{Heuristic Scoring}  \\
\midrule
\rowcolor{gray!10} \multicolumn{3}{l}{\textit{Answer Presence in Contexts}} \\
Exact Match       &  27.8\%  & 31.1\% \\
Fuzzy Match       & 4.4\% & 5.7\% \\
\midrule
\textbf{Total Found} & \textbf{32.2\% } & \textbf{ 36.8\%} \\
\bottomrule
\end{tabular}

\label{tab:retrieval_quality}
\end{table}
We observe an improvement of 11.9\% in exact match and 29.5\% in fuzzy match for our approach compared to baseline RAG, resulting in a total of 14.3\% improvement. This indicates that our mechanism helps preserve the correct answer spans.

\begin{table*}[!htpb]
\caption{Performance comparison of \myTitle\ utilizing generated captions (ChartGemma) versus plot digitization (DePlot).}
\label{tab:merged_context_comparison}
\centering
\begin{adjustbox}{width=0.85\linewidth}
\begin{tabular}{lcccccc}
\toprule
{\textbf{Question Type}} & \multicolumn{2}{c}{\textbf{ChartGemma with}} & \multicolumn{4}{c}{\textbf{DePlot with}} \\ \cmidrule{2-7}
 & \multicolumn{2}{c}{\textbf{\myTitle (with RAG)}} & \multicolumn{2}{c}{\textbf{\myTitle (with RAG)}} & \multicolumn{2}{c}{\textbf{\myTitle (without RAG)}} \\ \cmidrule{2-7}
 & \textbf{R1-F1} & \textbf{RL-F1} & \textbf{R1-F1} & \textbf{RL-F1} & \textbf{R1-F1} & \textbf{RL-F1} \\ \midrule
Closed-ended infinite (visual) & 0.416 & 0.416 & 0.493 & 0.493 & 0.486 & 0.486 \\
Closed-ended infinite (non-visual) & 0.662 & 0.659 & 0.654 & 0.650 & 0.663 & 0.661 \\
Closed-ended finite (binary visual) & 0.745 & 0.745 & 0.736 & 0.736 & 0.746 & 0.746 \\
Closed-ended finite (binary non-visual) & 0.788 & 0.788 & 0.806 & 0.806 & 0.811 & 0.811 \\
Closed-ended finite (non-binary visual) & 0.734 & 0.734 & 0.735 & 0.735 & 0.744 & 0.744 \\
Closed-ended finite (non-binary non-visual) & 0.751 & 0.751 & 0.762 & 0.762 & 0.757 & 0.757 \\
Unanswerable & 0.939 & 0.939 & 0.909 & 0.909 & 0.900 & 0.900 \\ \midrule
\textbf{Overall} & \textbf{0.719} & \textbf{0.719} & \textbf{0.728} & \textbf{0.727} & \textbf{0.730} & \textbf{0.729} \\
\bottomrule
\end{tabular}
\end{adjustbox}
\end{table*}
\section{Ablation Study}

To assess how effective our retrieval mechanism is, we compare the ground truth answers with the retrieved context blocks. Table \ref{tab:retrieval_quality}, represents the presence of the answer in contexts, providing a comparison between baseline RAG and our proposed RAG architecture consisting of heuristic re-ranking and query expansion. We use exact match and fuzzy match count to measure context relevance. The exact match measures the appearance of ground truth answers in a complete phrase with word boundaries within the RAG context, and  if answer tokens appear  $\geq 70\%$ somewhere in the contexts we call it fuzzy match.




Our primary focus was addressing the model's initial struggle with Closed-ended infinite (visual) questions, where purely chart understanding models typically fail. We conduct an ablation study analyzing the impact of synthetic captions generated by ChartGemma vs. plot digitization using DePlot with \myTitle. We speculate from Table \ref{tab:performance_comparison} that the model's inability to identify visual information in complex scientific charts is the cause of this failure. This is similar to research by Liu et al.~\cite{liu2023deplot} and Masry et al.~\cite{masry2022chartqa}, who observe that generalized chart models frequently fail to achieve the precision required for domain-specific scientific figures.

We also evaluate \myTitle with ChartGemma generated captions and plot digitization using DePlot in Table \ref{tab:merged_context_comparison}.  \myTitle with the ChartGemma configuration yields an R1-F1 and RL-F1 score of 0.719, representing a 2.7\% performance degradation relative to the \myTitle. We choose this approach to provide visual grounding for infinite visual questions through synthetic captions, but the accuracy of closed-ended infinite (visual) drops by 23.1\%. This demonstrates that captions through ChartGemma are not effective to provide the visual grounding required for scientific reasoning.

Inferring \myTitle with plot digitization using DePlot achieves an accuracy R1-F1 of 0.730 and RL-F1 of 0.729. Comparing this approach with the former ChartGemma setup, we observe an improvement of 18.5\% in the Closed-ended infinite (visual) question. From this we can conclude that plot digitization provides better visual grounding through precise number extraction.

Interestingly \myTitle(RAG) with the ChartGemma context configuration achieves the highest score on \textit {Unanswerable} questions (0.939). So we can conclude that ChartGemma's abstract captions are less specific than DePlot's data tables, making the model more conservative and less likely to force an answer from the text.

\section{Conclusion}
We present a resource-efficient architecture that enhances scientific reasoning through context-aware retrieval. Moreover, we address limitations of current state-of-the-art chart-understanding models in the scientific domain. Our results show that MLLMs struggle to answer closed-ended infinite visual questions that require precise visual grounding. To mitigate this limitation, our ablation study demonstrates that synthetic captions generated by chart-understanding models further reduce accuracy on visual questions, whereas plot digitization with DePlot improves accuracy. In future work, we will explore open-ended scientific visual question answering by integrating context-aware RAG with attention mechanisms.



\bibliographystyle{IEEEtran}
\bibliography{mybib}


\end{document}